\documentclass[11pt,a4paper]{article}
\usepackage{acl2017}
\usepackage{times}
\usepackage{latexsym}

\usepackage{times}
\usepackage{helvet}
\usepackage{courier}
\usepackage{amsmath}
\usepackage{bm}
\usepackage{float}
\usepackage{graphicx}
\usepackage{caption}
\usepackage{subfig}
\usepackage{multirow}

\aclfinalcopy
\begin{document}

\title{An Empirical Study of Adequate Vision Span for Attention-Based Neural Machine Translation}
\author{
Raphael Shu, Hideki Nakayama \\
\texttt{shu@nlab.ci.i.u-tokyo.ac.jp, nakayama@ci.i.u-tokyo.ac.jp} \\
\texttt{The University of Tokyo}
}

\maketitle
\begin{abstract}

Recently, the attention mechanism plays a key role to achieve high performance for Neural Machine Translation models. However, as it computes a score function for the encoder states in all positions at each decoding step, the attention model greatly increases the computational complexity. In this paper, we investigate the adequate vision span of attention models in the context of machine translation, by proposing a novel attention framework that is capable of reducing redundant score computation dynamically. The term ``vision span'' means a window of the encoder states considered by the attention model in one step. In our experiments, we found that the average window size of vision span can be reduced by over 50\% with modest loss in accuracy on English-Japanese and German-English translation tasks.% This results indicate that the conventional attention mechanism performs a significant amount of redundant computation.
\end{abstract}

\section{Introduction}

In recent years, recurrent neural networks have been successfully applied in machine translation. In many major language pairs, Neural Machine Translation (NMT) has already outperformed conventional Statistical Machine Translation (SMT) models \cite{luong2015addressing, wu2016google}.

NMT models are generally composed of an encoder and a decoder, which is also known as encoder-decoder framework \cite{sutskever2014sequence}. The encoder creates a vector representation of the input sentence, whereas the decoder generates the translation from this single vector. This simple encoder-decoder model suffers from a long backpropagation path; thus, adversely affected by long input sequences.

In recent NMT models, soft attention mechanism \cite{bahdanau2014neural} has been a key extension to ensure high performance. In each decoding step, the attention model computes alignment weights for all the encoder states. Then a context vector, which is a weighted summarization of the encoder states is computed and fed into the decoder as input. In contrast to the afore-mentioned simple encoder-decoder model, the attention mechanism can greatly shorten the backpropagation path.

Although the attention mechanism provides NMT models with a boost in performance, it also significantly increases the computational burden. As the attention model has to compute the alignment weights for all the encoder states in each step, the decoding process becomes time-consuming. Even worse, recent researches in NMT prefer to separate the texts into subwords \cite{rico2106bpe} or even characters \cite{chung2016char}, which means massive encoder states have to be considered in the attention model at each step, thereby resulting in increasing computational cost. On the other hand, the attention mechanism is becoming more complicated. For example, the NMT model with recurrent attention modeling \cite{yang2016neural} maintains a dynamic memory of attentions for every encoder states, which is updated in each decoding step. % NTN ?

% Old proposal introduction
%In this paper, we focus on reducing the amount of computations required by the attention mechanism. As online translation is now switching to NMT \cite{wu2016google}, reducing the complexity of attention mechanism can enable the translation systems to either process longer sequences in limited time or incorporate more expensive attention models.

\begin{figure}[tb]
  \centering
  \subfloat[]{\includegraphics[width = 0.45\textwidth]{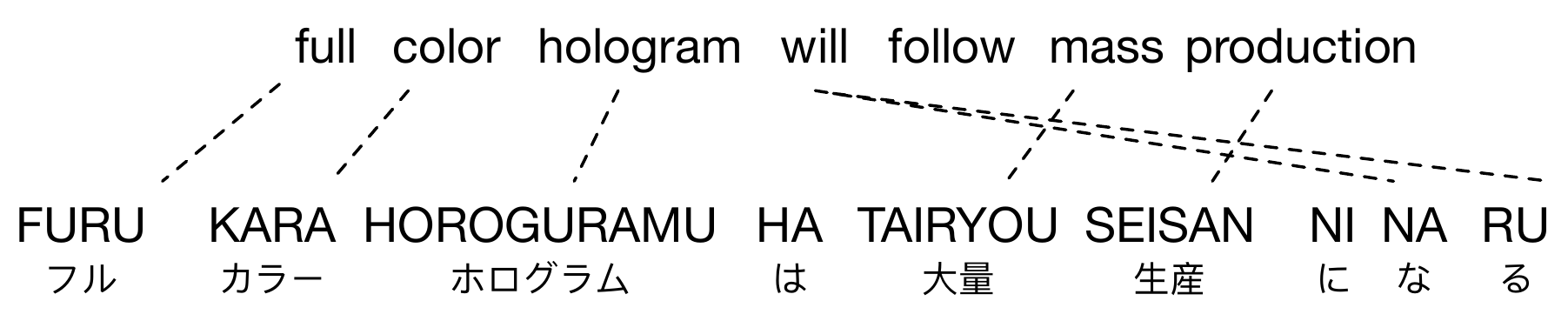}} \\
  \subfloat[]{\includegraphics[width = 0.45\textwidth]{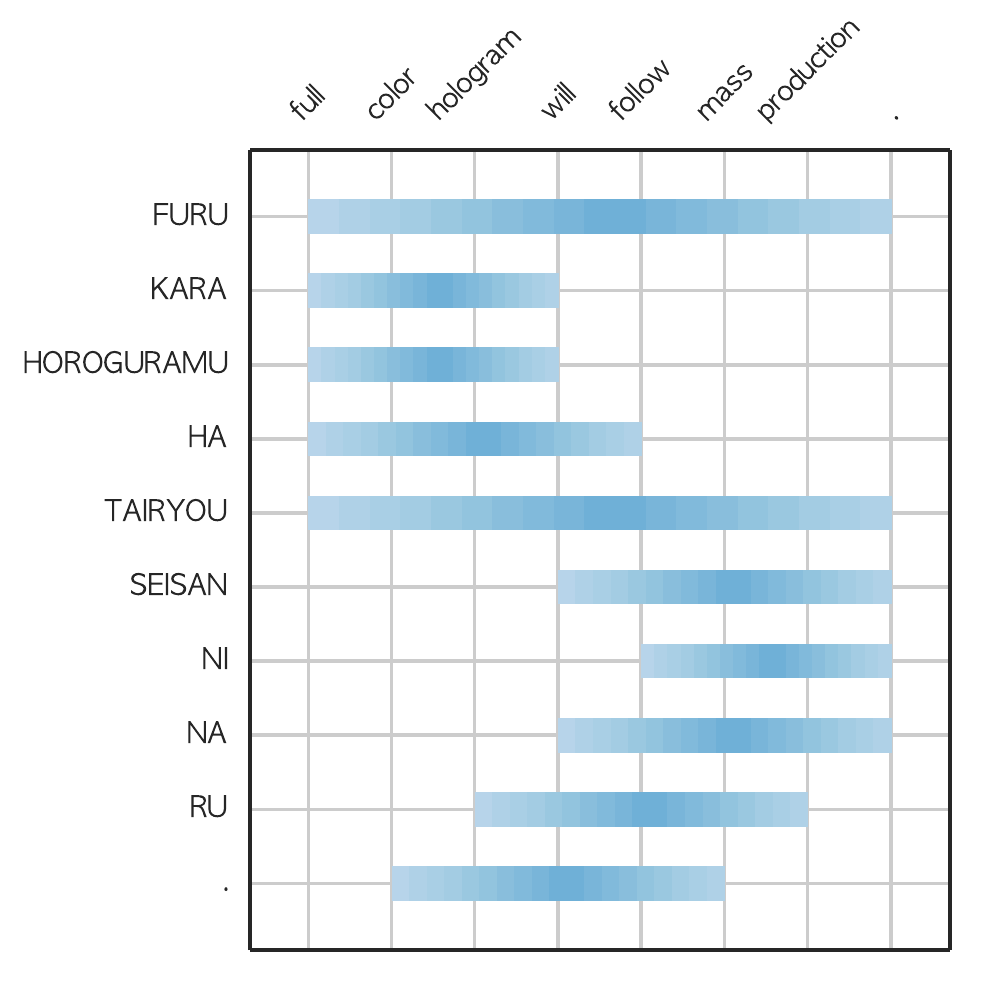}}
  \caption{(a) An example English-Japanese sentence pair with long-range reordering (b) The vision span predicted by the proposed Flexible Attention at each step in English-Japanese translation task}
  \label{figure:example_sent}
\end{figure}

% Important
In this paper, we study the adequate {\it vision span} in the context of machine translation. Here, the term ``vision span'' means a window of encoder states considered by the attention model in one step. We examine the minimum window size of an attention model have to consider in each step while maintaining the translation quality. For this purpose, we propose a novel attention framework which we refer to as Flexible Attention in this paper. The proposed attention framework tracks the center of attention in each decoding step, and predict an adequate vision span for the next step. In the test time, the encoder states outside of this range are omitted in the computation of score function.

% Intuition of proposed model
Our proposed attention framework is based on simple intuition. For most language pairs, the translations of words inside a phrase usually remain together. Even the translation of a small chunk usually does not mix with the translation of other words. Hence, information about distant words is basically unnecessary when translating locally. Therefore, we argue that computing the attention over all positions in each step is redundant. However, attending to distant positions remains important when dealing with long-range reordering. In Figure \ref{figure:example_sent}(a), we show an example sentence pair with long-range reordering, where the positions of the first three words have monotone alignments, but the fourth word is aligned to distant target positions. If we can predict whether the next word to translate is in a local position, the amount of redundant computation in the attention model can be safely reduced by controlling the window size of vision span dynamically. This motivated us to propose a flexible attention framework which predicts the minimum required vision span according to the context (See Figure \ref{figure:example_sent}(b)).

% Important
We evaluated our proposed Flexible Attention by comparing with the conventional attention mechanism, and Local Attention \cite{luong2015effective} which puts attention on a fixed-size window. We focus on comparing the minimum window size of vision span these models can achieve without hurting the performance too much. Note that as the window size determines the number of times the score function is evaluated, reducing the window size leads to the reduction of score computation. We select English-Japanese and German-English language pairs for evaluation as they consi    st of languages with different word orders, which means the attention model cannot simply look at a local range constantly and translate monotonically. Through empirical evaluation, we found with Flexible Attention, the average window size is reduced by 56\% for English-Japanese task and 64\% for German-English task, with modest loss of accuracy. The reduction rate also achieves 46\% for character-based NMT models.

Our contributions can be summarized as three folds:

\begin{enumerate}
  \item We empirically confirmed that the conventional attention mechanism performs a significant amount of redundant computation. Although attending globally is necessary when dealing with long-range reordering, a small vision span is sufficient when translating locally. The results may provide insights for future research on more efficient attention-based NMT models.

    \item The proposed Flexible Attention provides a general framework for reducing the amount of score computation according to the context, which can be combined with other expensive attention models of which computing for all positions in each step is costly.
    \item We found that reducing the amount of computation in the attention model can benefit the decoding speed on CPU, but not GPU.

% Mechanism itself
 % \item We proposed a novel attention model that controls the range of attention in a flexible manner in order to reduce the amount of the computational cost. We found that the amount of computation of attention model can be reduced by half with almost no harm to the accuracy.

\end{enumerate}

\section{Attention Mechanism in NMT}

% Another detailed introduction of neural machine translation

Although the network architectures of NMT models differ in various respects, they generally follow the encoder-decoder framework. In \citet{bahdanau2014neural}, a bidirectional recurrent neural network is used as the encoder, which accepts the embeddings of input words. The hidden states $\bm{\bar h_1}, ..., \bm{\bar h_{S}}$ of the encoder are then used in the decoding phase. Basically, the decoder is composed of a recurrent neural network (RNN). The decoder RNN computes the next state based on the embedding of the previously generated word, and a context vector given by the attention mechanism. Finally, the probabilities of output words in each time step are predicted based on the decoder states $\bm{h_1}, ..., \bm{h_N}$.

% Global attention
The soft attention mechanism \cite{karol2015draw} is introduced to NMT in \citet{bahdanau2014neural}, which computes a weighted summarization of all encoder states in each decoding step, to obtain the context vector:

\begin{equation}
    \bm{c}_t = \sum\limits_s a_t(s) \bm{\bar h_s} \: ,
\end{equation}
where $\bm{\bar h_s}$ is the $s$-th encoder state, $a_t(s)$ is the alignment weight of $\bm{\bar h_s}$ in decoding step $t$. The calculation of $a_t(s)$ is given by the softmax of the weight scores:

\begin{equation}
\label{equation:global_att}
\begin{split}
    a_t(s) = \frac{\exp(\mbox{score}(\bm{h_{t-1}}, \bm{\bar h_s))}}
    {\sum_{s^\prime}\exp(\mbox{score}(\bm{h_{t-1}}, \bm{\bar h_{s^\prime}}))}
\end{split} \: .
\end{equation}

The unnormalized weight scores are computed with a score function, defined as\footnote{In the original paper \cite{bahdanau2014neural}, the equation of the score function is a sum. Here, we use a concatenation in Equation \ref{equation:score} in order to align with \cite{luong2015effective}, which is an equivalent form of the original equation.}:

\begin{equation}
\label{equation:score}
    \mbox{score}(\bm{h_{t-1}}, \bm{\bar h_s}) = \bm{v_a}^\top \mbox{tanh} (\bm{W_a} [\bm{h_{t-1}};\bm{\bar h_s}]) \: ,
\end{equation}
where $\bm{v_a}$ and $\bm{W_a}$ are the parameters of the score function, $[\bm{h_{t-1}};\bm{\bar h_s}]$ is a concatenation of the decoder state in the previous step and an encoder state. Intuitively, the alignment weight indicates whether an encoder state is valuable for generating the next output word. Note that many discussions on alternative ways for computing the score function can be found in \citet{luong2015effective}.

% Coverage attention models

%Recently, many modifications to the attention mechanism are proposed in order to incorporate coverage information into NMT.   Basically those modifications are done by enhancing Equation \ref{equation:global_att} or Equation \ref{equation:score}. \cite{sankaran2016temporal} proposed a simple modification to explicitly memorize the attention temporally and penalize the weight of over-attended encoder states, which is referred to as Temporal Attention in their paper. The proposal of this model is motivated to track the coverage during decoding. Coverage embedding model \cite{mi2016coverage} was also proposed to solve the same problem, which employs coverage embeddings to capture the fertility of each words. The coverage embedding vector is loaded from the embedding table and updated in each decoding time step. Similarly, the model described in \cite{modeling2016tu} keeps track of the coverage information with a coverage vector. The coverage vector is fed into the score function in Equation \ref{equation:score}, and updated in each step.
%
%The model described in \cite{incorporating2016cohn} also modifies the score function in Equation \ref{equation:score}. Different from other related researches mentioned above, it incorporates the structural bias which is inspired from IBM models. This model does not only deal with fertility of words, but also considers positional bias, Markov condition and bilingual symmetry in the equation.

\section{Flexible Attention}

% Start to introduce the model formally
% TODO: revise this figure
\begin{figure*}
  \centering
  \includegraphics[width=0.9\textwidth]{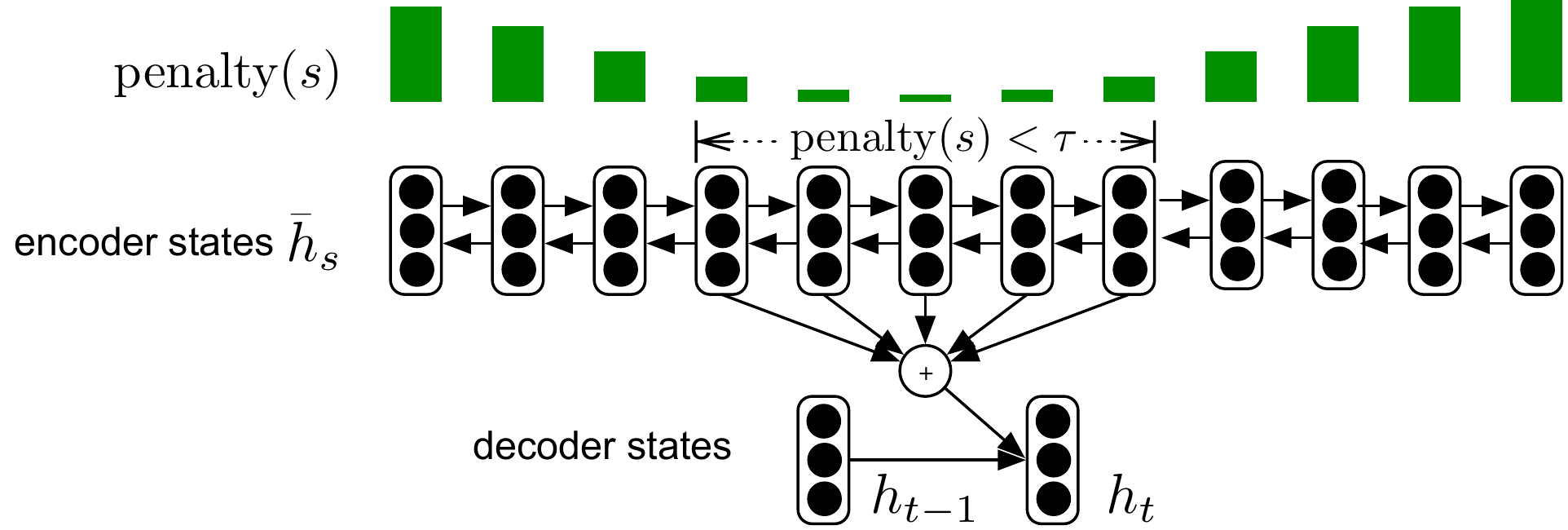}
  \caption{Illustration of the way that Flexible Attention reduces the window of vision span. In each decoding step, only a portion of the encoder states are selected by the position-based penalty function to compute the alignment weights. }
  \label{figure:flex_att}
\end{figure*}

In this section, we present our main idea for reducing the window size of vision span. In contrast to conventional attention models, we track the center of attention in each decoding step with

\begin{equation}
    p_t = \sum\limits_s a_t(s) \cdot s \: .
\end{equation}

The value of $p_t$ provides an approximate focus of attention in time step $t$. Then we penalize the alignment weights for the encoder states distant from $p_{t-1}$, which is the focus in the previous step. This is achieved by a position-based penalty function:

\begin{equation}
    \mbox{penalty}(s) = g(t) d(s, p_{t-1}) \: ,
    \label{equation:penalty}
\end{equation}
where $g(t)$ is a {\it sigmoid} function that adjusts the strength of the penalty dynamically based on the context in step $t$. $d(s, p_{t-1})$ provides the distance between position $s$ and the previous focus $p_{t-1}$, which is defined as:

\begin{equation}
    d(s, p_{t-1}) = \frac{1}{2 \sigma^2}(s - p_{t-1})^2 \: .
    \label{equation:distance}
\end{equation}

Hence, distant positions attract exponentially large penalties. The denominator $2\sigma^2$, which is a hyperparameter, controls the maximum of penalty when $g(t)$ outputs 1.

The position-based penalty function is finally integrated into the computation of the alignment weights as:
\begin{equation}
\label{equation:flex_att}
\begin{split}
    &a_t(s) =  \\
    &\frac{
    \exp(\mathrm{score}(\bm{h_{t-1}}, \bm{\bar h_s}) - \mbox{penalty}(s))
    }{
    \sum_{s^\prime}\exp(\mathrm{score}(\bm{h_{t-1}}, \bm{\bar h_{s^\prime}}) - \mbox{penalty}(s^\prime))},
\end{split}
\end{equation}
where the penalty function acts as a second score function that penalize encoder states only by their positions. When $g(t)$ outputs zero, the penalty function will have no effects on alignment weights. Note that the use of distance-based penalties here is similar in appearance to Local Attention (local-p) proposed in \citet{luong2015effective}. The difference is that Local Attention predicts the center of attention in each step and attends to a fixed window. Further discussion will be given later in Section \ref{section:related}.

In this paper, the strength function $g(t)$ in Equation \ref{equation:penalty} is defined as:

\begin{equation}
\begin{split}
g(t) = \mathrm{sigmoid} (\bm{v_g}^\top \mathrm{tanh}(\bm{W_g} [\bm{h_{t-1}};\bm{i_t}]) + b_g)
\end{split},
\label{eqn:g_prime}
\end{equation}
where $\bm{v_g}$, $\bm{W_g}$ and $b_g$ are parameters. We refer to this attention framework as Flexible Attention in this paper, as the window of effective attention is adjusted by $g(t)$ in according to the context.

Intuitively, when the model is translating inside a phrase, the alignment weights for distant positions can be safely penalized by letting $g(t)$ output a high value. If the next word is expected to be translated from a new phrase, $g(t)$ shall output a low value to allow attending to any position. Actually, the selected output word in the previous step can greatly influence this decision, as selection of the output word can determine whether the translation of a phrase is complete. Therefore, the embedding of the feedback word $\bm{i_t}$ is put into the equation.

\subsection{Reducing Window Size of Vision Span}
% TODO: why only in test time?

As we can see from Equation \ref{equation:flex_att}, if a position is heavily penalized, then it will be assigned a low attention probability regardless of the value of the score function. In the test time, we can set a threshold $\tau$, and only compute the score function for positions with penalties lower than $\tau$. Figure \ref{figure:flex_att} provides an illustration of the selection process. The selected range can be obtained by solving $penalty(s) < \tau$, which gives $s \in \bigl( \: p_{t-1} - \sigma \sqrt{2 \tau / g(t) }, \:\: p_{t-1} + \sigma \sqrt{2 \tau / g(t)} \: \bigl) $.

Because the strength term $g(t)$ in Equation \ref{equation:penalty} only needs to be computed once in each step, the computational cost of the penalty function does not increase as the input length increases. By utilizing the penalty values to omit computation of the score function, the totally computational cost can be reduced.

Although a low threshold would lead to further reduction of the window size of vision span, the performance degrades as information from the source side will be greatly limited. In practice, we can find a good threshold to balance the tradeoff of performance and computational cost on a validation dataset.

\subsection{Fine-tuning for Better Performance}

In order to further narrow down the vision span, we want $g(t)$ to output a large value to clearly differentiate valuable encoder states from other states encoder based on their positions. Thus, we can further fine-tune our model to encourage it to decode using larger penalties with the following loss function:

\begin{equation}
J = \sum^D_{i=1} - \log p(y^{(i)}|x^{(i)}) - \beta \frac{1}{T} \sum\limits_{t=1}^{T} g(t)^{(i)},
\end{equation}
where $\beta$ is a hyperparameter to control the balance of cross-entropy and the average strength of penalty. In our experiments, we tested $\beta$ among $(0.1, 0.001, 0.0001)$ on a development data and found that setting $\beta$ to $0.1$ and fine-tuning for one epoch works well. If we train the model with this loss function from the beginning, as the right part of the loss function is easier to be optimized, the value of $g(t)$ saturates quickly, which slows down the training process.

\section{Related Work}
\label{section:related}

To the best of our knowledge, only a limited number of related studies aimed to reduce the computational cost of the attention mechanism. Local Attention, which was proposed in \citet{luong2015effective}, limited the range of attention to a fixed window size. In Local Attention (local-p), the center of attention $p_t$ is {\it predicted} in each time step $t$:

\begin{equation}
    p_t = S \cdot \operatorname{sigmoid} (\bm{v_p}^\top \operatorname{tanh}(\bm{W_p} \bm{h_t})) \: ,
\end{equation}
where $S$ is the length of the input sequence. Finally, the alignment weights are computed by:

\begin{equation}
\label{equation:local_att}
\begin{split}
    a^\prime_t(s) &= a_t(s) \exp (-\frac{(s-p_t)^2}{2 \sigma^2})\\
    &= \frac{\small
       exp(\mbox{score}(\bm{h_{t-1}}, \bm{\bar h_s))}
    }{\small
       \sum_{s^\prime}exp(\mbox{score}(\bm{h_{t-1}}, \bm{\bar h_{s^\prime}}))
    } \exp (-\frac{(s-p_t)^2}{2 \sigma^2}),
\end{split}
\end{equation}
where $\sigma$ is a hyperparameter determined by $\sigma = D / 2$, where $D$ is a half of the window size. Local Attention only computes attention within the window $[p_t-D,p_t+D]$. In their work, the hyperparameter $D$ is empirically set to $D=10$ for the English-German translation task, which means a window of 21 words.

Our proposed attention model differs from Local Attention in two key points: (1) our proposed attention model does not predict the fixiation of attention but tracks it in each step (2) the position-based penalty in our attention model is adjusted flexibly rather than remaining fixed. Note that in Equation \ref{equation:local_att} of Local Attention, the penalty term is applied outside the softmax function. In contrast, we integrate the penalty term with the score function (Eq. \ref{equation:flex_att}), such that the final probabilities still add up to 1.

Recently, a ``cheap'' linear model \cite{de2016cheap} is proposed to replace the attention mechanism with a low-complexity function. This cheap linear attention mechanism achieves an accuracy in the middle of Global Attention and a non-attention model on a question-answering dataset. This approach can be considered as another interesting way to balance the performance and computational complexity in sequence-generation tasks.

% Other works?

\section{Experiments}
% TODO: Check this part again
In this section, we focus on evaluating our proposed attention models by measuring the minimum average window size of vision span it can achieve with a modest performance loss\footnote{In our experiments, we try to limit the performance loss to be lower than 0.5 development BLEU. As the threshold $\tau$ is selected using a development corpus, the performance on test data is not ensured.}. In detail, we measure the average number of the encoder states considered when computing the score function in Equation \ref{equation:score}. Note that as we decode using Beam Search algorithm \cite{sutskever2014sequence} , the value of window size is further averaged over the number of hypotheses considered in each step. For the conventional attention mechanism, as all positions have to be considered in each step, the average window size equals to the average sentence length of the testing data. Following \citet{luong2015effective}, we refer to the conventional attention mechanism as Global Attention in experiments.

% --->
\subsection{Experimental Settings}

\begin{table*}[t]
  \begin{center}
    \begin{tabular}{r|c|c|c|c|c} \hline \hline
      \multirow{2}{*}{\small{Model}} & \multicolumn{3}{c|}{\small{English-Japanese}} & \multicolumn{2}{c}{\small{German-English}} \\
      \cline{2-6}
       & \small{window (words)} &  \small{BLEU(\%)} &  \small{RIBES} & \small{window (words)} & \small{BLEU(\%)}  \\
%      \hline
%      \small{Tree-to-sequence model } & 81.42 & 34.65 \\
      \hline
      \small{Global Attention baseline} & 24.4 & 34.87 & 0.810 & 20.7 & 20.62  \\
      \small{Local Attention baseline} & 18.4 & 34.52 & 0.809 & 15.7 & 21.09 \\
      \hline
      \small{Flexible Attention ($\tau$=$\infty$)} & 24.4 & 35.01 & 0.814 & 20.7 & 21.31 \\
      \small{Flexible Attention ($\tau$=1.2)} & 16.4 & 34.90 & 0.812 & 7.8 & 21.11 \\
      \small{+ fine-tuning ($\tau$=1.2)} & {\bf 10.7} & 34.78 & 0.807 & {\bf 7.4} & 20.79\\
%      \small{Variation 1 (dynamic window size)} & 81.10 & \bf{35.16} \\
%      \small{Variation 2 (phrase-level attention)} & 80.95 & 34.54 \\
       \hline \hline
    \end{tabular}
    \caption{Evaluation results on English-Japanese and German-English translation task. This table provides a comparison of the minimum window size of vision span the models can achieve with a modest loss of accuracy. }
    \label{table:enja}
  \end{center}
\end{table*}

We evaluate our models on English-Japanese and German-English translation task. As translating these language pairs requires long-range reordering, the proposed Flexible Attention has to correctly predict when the reordering happens and look at distant positions when necessary. The training data of En-Ja task is based on ASPEC parallel corpus \cite{NAKAZAWA16.621}, which contains 3M sentence pairs, whereas the test data contains 1812 sentences, which have 24.4 words on average. We select 1.5M sentence pairs according to the automatically calculated matching scores, which are provided along with the ASPEC corpus. For De-En task, we use the WMT'15 training data consisting of 4.5M sentence pairs. The WMT'15 test data (newstest2015) contains 2169 pairs, which have 20.7 words on average.

We preprocess the En-Ja corpus with ``tokenizer.perl'' for English side, and Kytea tokenizer \cite{kytea} for Japanese side. The preprocessing procedure for De-En corpus is similar to \citet{li2014dcu}, except we did not filter sentence pairs with language detection.

The vocabulary size are cropped to 80k and 40k for En-Ja NMT models, whereas 50k for De-En NMT models. The OOV words are replaced with a ``UNK'' symbol.  Long sentences with more than 50 words on either the source or target side are removed from the training set, resulting in 1.3M and 3.8M training pairs for En-Ja and De-En task respectively. We use mini-batch in our training procedure, where each batch contains 64 data samples. All sentence pairs are firstly sorted according to their length before we group them into batches. After which, the order of the mini-batches is shuffled.

We adopt the network architecture described in \citet{bahdanau2014neural} and set it as our baseline model. The size of word embeddings is 1000 for both languages. For the encoder, we use a bi-directional RNN composed of two LSTMs with 1000 units. 　For the decoder, we use a one-layer LSTM with 1000 units, where the input in each step is a concatenated vector of the embedding of the previous output $i_t$ and the context vector $c_t$ given by attention mechanism. Before the final softmax layer, we insert a fully-connected layer with 600 units to reduce the number of connections in the output layer.

% Important
For our proposed models, we empirically select $\sigma$ in Equation \ref{equation:distance} from $(\frac{3}{2}, \frac{10}{2}, \frac{15}{2}, \frac{20}{2})$ on a development corpus. In our experiments, we found the attention models give the best trade-off between the window size and accuracy when $\sigma = 1.5$. Note that the value of $\sigma$ only determines the maximum of penalty when $g(t)$ outputs $1$, but does not results in a fixed window size.

% This choice of $\sigma$ indicates that Flexible Attention will attend to a extremely limited range of encoder states when $g(t)$ outputs $1$.

The NMT models are trained using Adam optimizer \cite{kingma2014adam} with an initial learning rate of 0.0001. We train the model for six epochs and start to halve the learning rate from the beginning of the fourth epoch. The maximum norm of the gradients is clipped to 3. Final parameters are selected by the smoothed BLEU \cite{smoothed_bleu} on validation set. During test time, we use beam search with a beam size of 20.

In En-Ja task, we evaluate our implemented NMT models with BLEU and RIBES \cite{isozaki2010automatic}, in order to align with other researches on the same dataset. The results are reported following standard post-processing procedures\footnote{We report the scores using Kytea tokenizer. The post-processing procedure for evaluation is described in \url{http://lotus.kuee.kyoto-u.ac.jp/WAT/evaluation/}}. For De-En task, we report {\it tokenized} BLEU \footnote{The scores are produced by tokenizing with ``tokenizer.perl'' and evaluating with ``multi-bleu.perl''.}.

% \footnote{Described in http://lotus.kuee.kyoto-u.ac.jp/WAT/ \\ evaluation/automatic\_evaluation\_systems/automaticEvaluationJA.html}.

\subsection{Evaluations of Flexible Attention}

We evaluate the attention models to determine the minimum window size they can achieve with a modest loss of accuracy (0.5 development BLEU) compared to Flexible Attention with $\tau = \infty$. The results we obtained are summarized in Table \ref{table:enja}. The scores of Global Attention (conventional attention model) and Local Attention \cite{luong2015effective} are listed for comparison. For Local Attention, we found a window size of 21 ($D=10$) gives the best performance for En-Ja and De-En tasks. In this setting, Local Attention achieves an average window of 18.4 words in En-Ja task and 15.7 words in De-En task, as some sentences in the test corpus have fewer than 21 words.

% TODO: select a threshold
For Flexible Attention, we search a good $\tau$ among ($0.8, 1.0, 1.2, 1.4, 1.6$) on a development corpus so that the development BLEU(\%) does not degrade more than $0.5$ compared to $\tau = \infty $. Finally, $\tau = 1.2$ is selected for both language pairs in our experiments.

We can see from the results that Flexible Attention can achieve comparable scores even consider only half of the encoder states in each step. After fine-tuning, our proposed attention model further reduces 56\% of the vision span for En-Ja task and 64\% for De-En task. The high reduction rate confirms that the conventional attention model performs massive redundant computation. With Flexible Attention, redundant score computation can be efficiently cut down according to the context. Interestingly, the NMT models using Flexible Attention without the threshold improves the translation accuracy by a small margin, which may indicates that the quality of attention is improved.

\subsection{Trade-off between Window Size and Accuracy}

\begin{figure}[tb]
  \centering
  \includegraphics[width=0.5\textwidth]{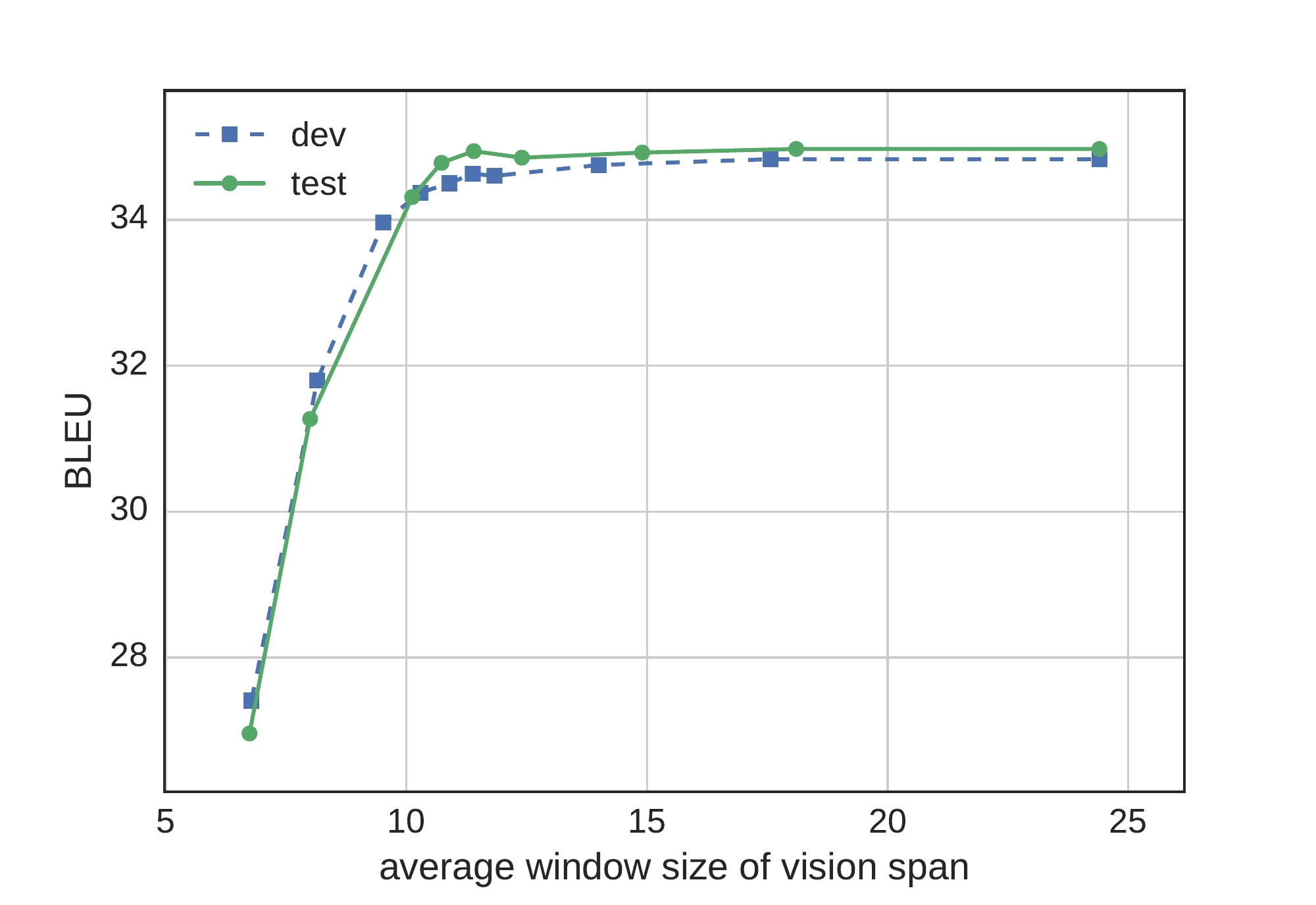}
  \caption{Trade off between window size and performance on the development and test data of English-Japanese tanslation task}
  \label{figure:tradeoff}
\end{figure}

% Important
In order to figure out the relation between accuracy and the window size of vision span, we plot out the curve of the trade-off between BLEU score and average window size on En-Ja task, which is shown in Figure \ref{figure:tradeoff}.

The data points are collected by testing different thresholds \footnote{In detail, the data points in the plot is based on the thresholds in $(0.3, 0.5, 0.8, 1.0, 1.2, 1.4, 1.6, 5.0, 8.0, 999)$.} with the fine-tuned Flexible Attention model. Interestingly, the NMT model with our proposed Flexible Attention suffers almost no loss in accuracy even the computations are reduced by half. Further trails to reduce the window size beneath 10 words will result in drastically degradation in performance.

\subsection{Effects on Character-level Attention}

\begin{table}[htb]
  \begin{center}

    \begin{tabular}{r|c|c|c} \hline \hline
      \small{Model} & \small{window} &  \small{BLEU} &  \small{RIBES}  \\
      \hline
      \small{Global Attention baseline} & 144.9 & 26.18 & 0.767   \\
      \hline
      \small{Flexible Attention ($\tau$=$\infty$)} & 144.9 & 26.68 & 0.763 \\
      \small{Flexible Attention ($\tau$=1.0)} & 80.4 & 26.18 & 0.757\\
      \small{+ fine-tuning ($\tau$=1.0)} & {\bf 77.4} & 26.23 & 0.757\\
      \hline \hline
    \end{tabular}
    \caption{Evaluation results with character-based English-Japanese NMT models}
    \label{table:enja_char}
  \end{center}
\end{table}

In order to examine the effects of Flexible Attention on extremely long character-level inputs, we also conducted experiments on character-based NMT models. We adopt the same network architecture as word-based models in the English-Japanese task, unless the sentences in both sides are tokenized into characters. We keep 100 most frequent types of character for the English side and 3000 types for the Japanese side. The embedding size is set to 100 for both sides. In order to train the models faster, all LSTMs in this experiment have 500 hidden units. The character-based models are trained 20 epochs with Adam optimizer with an initial learning rate of 0.0001. The learning rate begins to halve from 18-th epoch. After fine-tuning the model with the same hyperparameter ($\beta = 0.1$), we selected the threshold to be $\tau=1.0$ in the same manner as the word-level experiment. We did not evaluate Local Attention in this experiment as selecting a proper fixed window size is time-consuming when the length of input sequence is extremely long.

The experimental results of character-based models are summarized in Table \ref{table:enja_char}. Note that although the performance of character-based models can not compete with word-based model, the focus of this experiment is to examine the effects in terms of the reduction of the window size of vision span. For this dataset, the character-level tokenization will increase the length of input sequences by 6x on average. In this setting, the fine-tuned Flexible Attention model can achieve a reduction rate of 46\% of the vision span. The results indicate that Flexible Attention can automatically adapt to the type of training data and learn to control the strength of penalty properly.

\subsection{Impact on Real Decoding Speed}

In this section, we examine the impact of the reduction of score computation in terms of real decoding speed. We compare the fine-tuned Flexible Attention ($\tau=1.0$) with the conventional Global Attention on the English-Japanese dataset with character-level tokenization. \footnote{For word-level tasks, as the ``giant'' output layer has large impact on decoding time, we selected the character-level task to measure real decoding speed.}

We decode 5,000 sentences in the dataset and report the averaged decoding time on both GPU \footnote{NVIDIA GeForce GTX TITAN X.} and CPU \footnote{Intel Core\textsuperscript{TM} i7-5960X CPU @ 3.00GHz, single core. The implementation uses Theano with openblas as numpy backend.}. For each sequence, the dot production with $\bm{{\bar h}_s}$ in the score function (Eq. \ref{equation:score}) is pre-computed and cached before decoding. As different attention models will produce different numbers of output tokens for a same input, that the decoding time will be influenced by different computation steps of the decoder LSTM. In order to fairly compare the decoding time, we force the decoder to use the tokens in the reference as feedbacks. Thus, the number of decoding steps remains the same for both models.

\begin{table}[h]
  \begin{center}

    \begin{tabular}{r|c|c} \hline \hline
      \small{Model} & \small{avg. time (GPU)} & \small{avg. time (CPU)} \\
      \hline
      \small{Global Attention} & 123ms & 751ms \\
      \small{Flexible Attention} & 136ms & 677ms \\
      \hline \hline
    \end{tabular}
    \caption{Average decoding time for one sentence on the English-Japanese dataset with character-level tokenization}
    \label{table:spd}
  \end{center}
\end{table}

As shown in Table \ref{table:spd}, reducing the amount of computation in attention model is shown to benefit the decoding speed on CPU. However, applying Flexible Attention slows down the decoding speed on GPU. This is potentially due to the overhead of computing the strength of penalty in Equation \ref{eqn:g_prime}.

For the CPU-based decoding, after profiling our Theano code, we found that the output layer is the main bottleneck, which accounts for 58\% of the computation time. In a recent paper \cite{l2016vocabulary}, the authors show that CPU decoding time can be reduced by 90\% by reducing the computation of the output layer, resulting in just over 140ms per sentence. Our proposed attention model has the potential to be combined with their method to further reduce the decoding time on CPU.  

As the score function we use in this paper has relatively low computation cost, the difference of real decoding speed is expected to be enlarged with more complicated attention models, such as Recurrent Attention \cite{yang2016neural} and Neural Tensor Network \cite{NIPS2013_5028}.

\subsection{Qualitative Analysis of Flexible Attention}
\label{section:qa}

%TODO: fix
In order to inspect the behaviour of the penalty function in Equation \ref{equation:flex_att}, we let the NMT model translate the sentence in Figure \ref{figure:example_sent}(a) and record the word positions the attention model considers in each step. The vision span predicted by Flexible Attention is visionized in Figure \ref{figure:example_sent}(b).

We can see that the value of $g(t)$ changes dynamically in different context, resulting in different vision span in each step. In the most of the time, the attention is constrained in a local span when translating inside phrases. When emitting the fifth word ``TAIRYOU'', as the reordering occurs, the attention model looks globally to find the next word to translate. Analyzing the vision spans predicted by Flexible Attention in De-En task also shows similar result that the model only attends to a large span occasionally. The qualitative analysis of Flexible Attention confirms our hypothesis that attending globally in each step is redundant for machine translation. More visualizations can be found in the supplementary material.

% During the translation of ``thyrotoxicosis'', which has one-to-many alignment, a high strength of penalty is given by $g(t)$ to lock the attention to the same word. After finishing the translation of ``thyrotoxicosis'', the attention model searches for the next word to translate globally.

\section{Conclusion}

In this paper, we proposed a novel attention framework that is capable of reducing the window size of attention dynamically according to the context. In our experiments, we found the proposed model can safely reduce the window size by 56\% for English-Japanese and 64\% German-English task on average. For character-based models, our proposed Flexible Attention can also achieve a reduction rate of 46\%.

In qualitative analysis, we found that Flexible Attention only needs to put attention on a large window occasionally, especially when long-range reordering is required. The results confirm the existence of massive redundant computation in the conventional attention mechanism. By cutting down unnecessary computation, NMT models can translate extremely long sequence efficiently or incorporate more expensive score functions.

\section*{Acknowledgement}

This work is supported by JSPS KAKENHI Grant Number 16H05872. 

\bigskip

\bibliography{mybib}

\begin{thebibliography}{}
\expandafter\ifx\csname natexlab\endcsname\relax\def\natexlab#1{#1}\fi

\bibitem[{Bahdanau et~al.(2014)Bahdanau, Cho, and Bengio}]{bahdanau2014neural}
Dzmitry Bahdanau, Kyunghyun Cho, and Yoshua Bengio. 2014.
\newblock Neural machine translation by jointly learning to align and
  translate.
\newblock {\em arXiv preprint arXiv:1409.0473\/} .

\bibitem[{Chung et~al.(2016)Chung, Cho, and Bengio}]{chung2016char}
Junyoung Chung, Kyunghyun Cho, and Yoshua Bengio. 2016.
\newblock \href{https://doi.org/10.18653/v1/P16-1160}{A character-level decoder
  without explicit segmentation for neural machine translation}.
\newblock In {\em ACL\/}. pages 1693--1703.
\newblock
  \href{https://doi.org/10.18653/v1/P16-1160}{https://doi.org/10.18653/v1/P16-1160}.

\bibitem[{de~Br{\'e}bisson and Vincent(2016)}]{de2016cheap}
Alexandre de~Br{\'e}bisson and Pascal Vincent. 2016.
\newblock A cheap linear attention mechanism with fast lookups and fixed-size
  representations.
\newblock {\em arXiv preprint arXiv:1609.05866\/} .

\bibitem[{Isozaki et~al.(2010)Isozaki, Hirao, Duh, Sudoh, and
  Tsukada}]{isozaki2010automatic}
Hideki Isozaki, Tsutomu Hirao, Kevin Duh, Katsuhito Sudoh, and Hajime Tsukada.
  2010.
\newblock Automatic evaluation of translation quality for distant language
  pairs.
\newblock In {\em EMNLP\/}. pages 944--952.

\bibitem[{Kingma and Ba(2014)}]{kingma2014adam}
Diederik Kingma and Jimmy Ba. 2014.
\newblock Adam: A method for stochastic optimization.
\newblock {\em arXiv preprint arXiv:1412.6980\/} .

\bibitem[{L'Hostis et~al.(2016)L'Hostis, Grangier, and Auli}]{l2016vocabulary}
Gurvan L'Hostis, David Grangier, and Michael Auli. 2016.
\newblock Vocabulary selection strategies for neural machine translation.
\newblock {\em arXiv preprint arXiv:1610.00072\/} .

\bibitem[{Li et~al.(2014)Li, Wu, Va{\i}llo, Xie, Way, and Liu}]{li2014dcu}
Liangyou Li, Xiaofeng Wu, Santiago~Cort{\'e}s Va{\i}llo, Jun Xie, Andy Way, and
  Qun Liu. 2014.
\newblock The dcu-ictcas mt system at wmt 2014 on german-english translation
  task.
\newblock In {\em Proceedings of the Ninth Workshop on Statistical Machine
  Translation\/}. pages 136--141.

\bibitem[{Lin and Och(2004)}]{smoothed_bleu}
Chin-Yew Lin and Franz~Josef Och. 2004.
\newblock \href{https://doi.org/10.3115/1218955.1219032}{Automatic evaluation
  of machine translation quality using longest common subsequence and
  skip-bigram statistics}.
\newblock In {\em ACL\/}.
\newblock
  \href{https://doi.org/10.3115/1218955.1219032}{https://doi.org/10.3115/1218955.1219032}.

\bibitem[{Luong et~al.(2015{\natexlab{a}})Luong, Pham, and
  Manning}]{luong2015effective}
Thang Luong, Hieu Pham, and D.~Christopher Manning. 2015{\natexlab{a}}.
\newblock \href{https://doi.org/10.18653/v1/D15-1166}{Effective approaches to
  attention-based neural machine translation}.
\newblock In {\em EMNLP\/}. pages 1412--1421.
\newblock
  \href{https://doi.org/10.18653/v1/D15-1166}{https://doi.org/10.18653/v1/D15-1166}.

\bibitem[{Luong et~al.(2015{\natexlab{b}})Luong, Sutskever, Le, Vinyals, and
  Zaremba}]{luong2015addressing}
Thang Luong, Ilya Sutskever, Quoc Le, Oriol Vinyals, and Wojciech Zaremba.
  2015{\natexlab{b}}.
\newblock Addressing the rare word problem in neural machine translation.
\newblock In {\em ACL\/}. pages 11--19.

\bibitem[{Nakazawa et~al.(2016)Nakazawa, Yaguchi, Uchimoto, Utiyama, Sumita,
  Kurohashi, and Isahara}]{NAKAZAWA16.621}
Toshiaki Nakazawa, Manabu Yaguchi, Kiyotaka Uchimoto, Masao Utiyama, Eiichiro
  Sumita, Sadao Kurohashi, and Hitoshi Isahara. 2016.
\newblock Aspec: Asian scientific paper excerpt corpus.
\newblock In {\em Proceedings of the Ninth International Conference on Language
  Resources and Evaluation (LREC 2016)\/}. pages 2204--2208.

\bibitem[{Neubig et~al.(2011)Neubig, Nakata, and Mori}]{kytea}
Graham Neubig, Yosuke Nakata, and Shinsuke Mori. 2011.
\newblock Pointwise prediction for robust, adaptable japanese morphological
  analysis.
\newblock In {\em ACL\/}. pages 529--533.

\bibitem[{Sennrich et~al.(2016)Sennrich, Haddow, and Birch}]{rico2106bpe}
Rico Sennrich, Barry Haddow, and Alexandra Birch. 2016.
\newblock \href{https://doi.org/10.18653/v1/P16-1162}{Neural machine
  translation of rare words with subword units}.
\newblock In {\em ACL\/}. pages 1715--1725.
\newblock
  \href{https://doi.org/10.18653/v1/P16-1162}{https://doi.org/10.18653/v1/P16-1162}.

\bibitem[{Socher et~al.(2013)Socher, Chen, Manning, and Ng}]{NIPS2013_5028}
Richard Socher, Danqi Chen, Christopher~D Manning, and Andrew Ng. 2013.
\newblock Reasoning with neural tensor networks for knowledge base completion.
\newblock In {\em NIPS\/}, pages 926--934.

\bibitem[{Sutskever et~al.(2014)Sutskever, Vinyals, and
  Le}]{sutskever2014sequence}
Ilya Sutskever, Oriol Vinyals, and Quoc~VV Le. 2014.
\newblock Sequence to sequence learning with neural networks.
\newblock In {\em NIPS\/}. pages 3104--3112.

\bibitem[{Wu et~al.(2016)Wu, Schuster, Chen, Le, Norouzi, Macherey, Krikun,
  Cao, Gao, Macherey et~al.}]{wu2016google}
Yonghui Wu, Mike Schuster, Zhifeng Chen, Quoc~V Le, Mohammad Norouzi, Wolfgang
  Macherey, Maxim Krikun, Yuan Cao, Qin Gao, Klaus Macherey, et~al. 2016.
\newblock Google's neural machine translation system: Bridging the gap between
  human and machine translation.
\newblock {\em arXiv preprint arXiv:1609.08144\/} .

\bibitem[{Yang et~al.(2016)Yang, Hu, Deng, Dyer, and Smola}]{yang2016neural}
Zichao Yang, Zhiting Hu, Yuntian Deng, Chris Dyer, and Alex Smola. 2016.
\newblock Neural machine translation with recurrent attention modeling.
\newblock {\em arXiv preprint arXiv:1607.05108\/} .

\end{thebibliography}
\bibliographystyle{acl_natbib}

\appendix

\section{Supplemental Material}
\label{sec:supplemental}

In order to give a better intuition for the conclusion in \ref{section:qa}, we show a visionization of the vision spans our proposed Flexible Attention predicted on the De-En development corpus. The NMT model can translate well without attending to all positions in each step. In contrast the conventional attention models, Flexible Attention only attends to a large window occasionally. 

%% The code for reproducing these results is available in \url{https://this_url_will_be_changed}.

\begin{figure*}[tb]
  \centering
  \subfloat[]{\includegraphics[width = 0.5\textwidth]{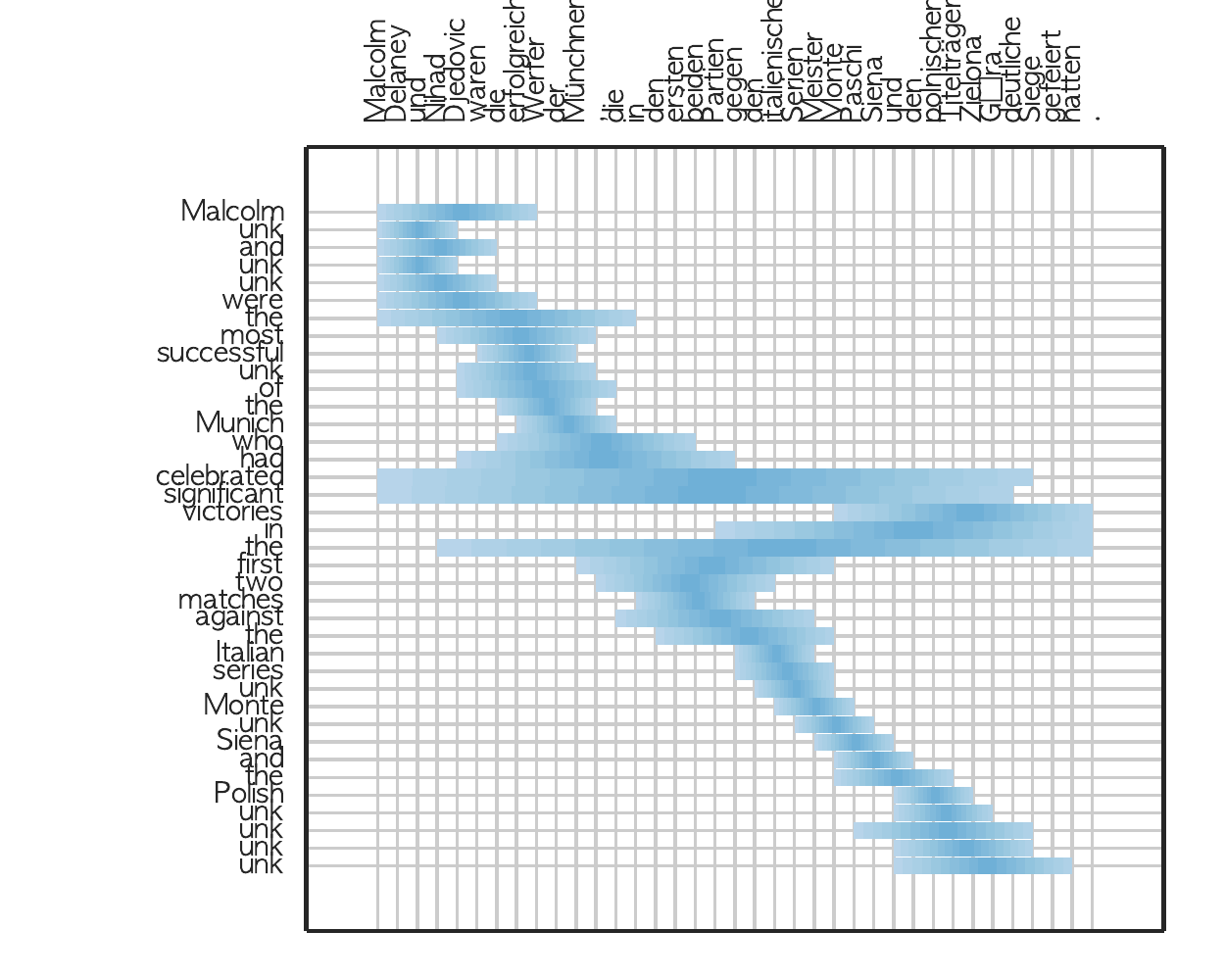}}
  \subfloat[]{\includegraphics[width = 0.5\textwidth]{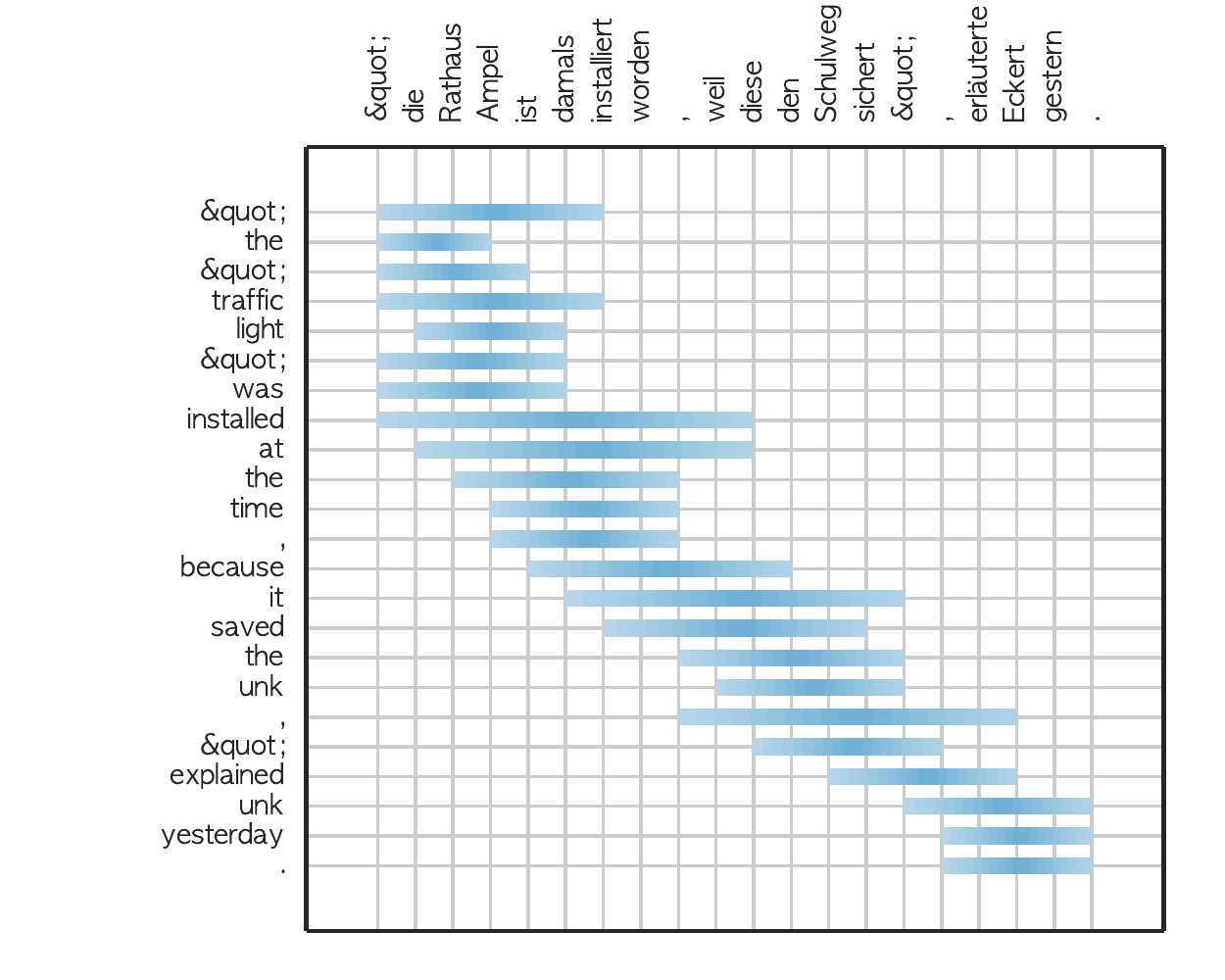}} \\
  \subfloat[]{\includegraphics[width = 0.5\textwidth]{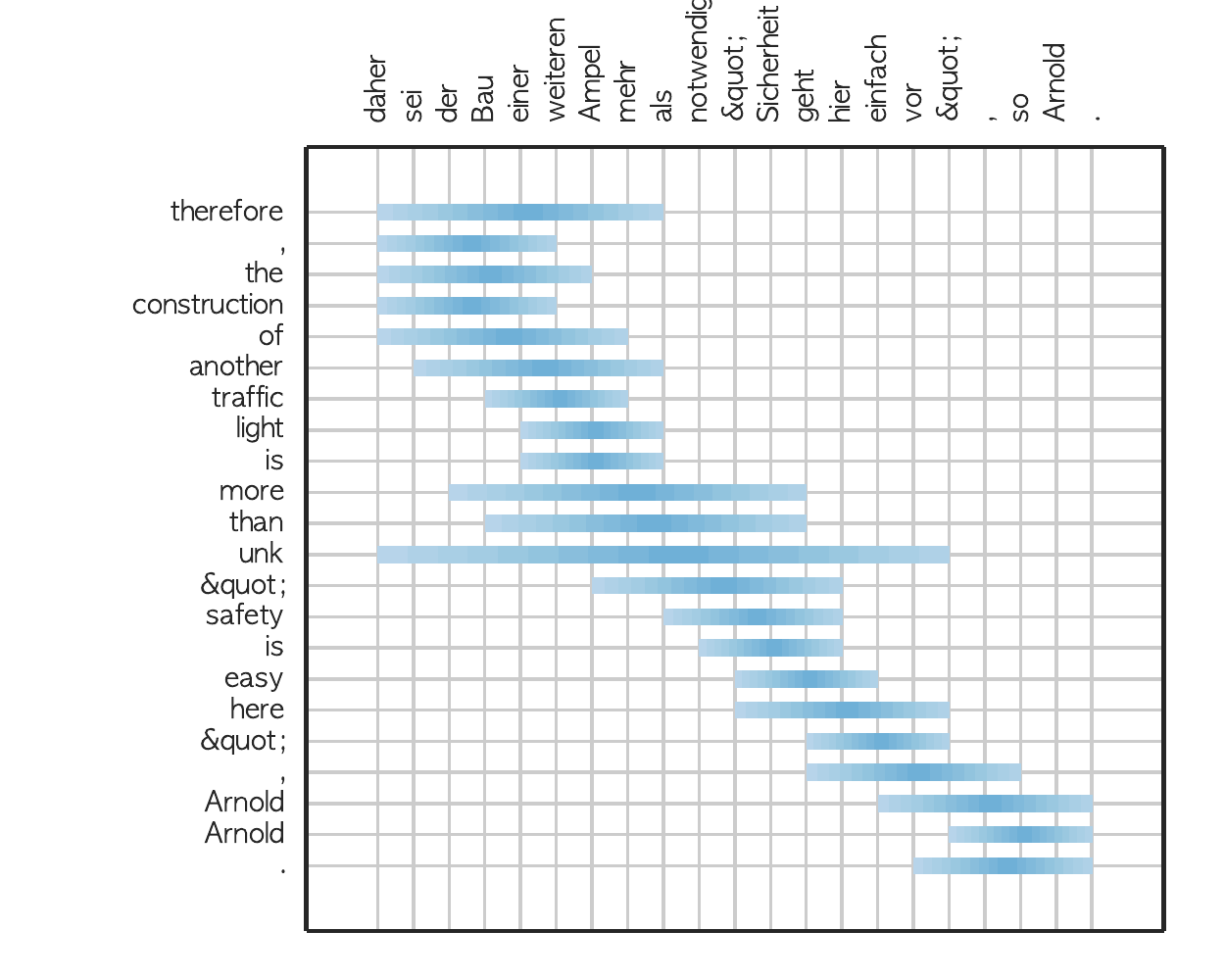}}
  \subfloat[]{\includegraphics[width = 0.5\textwidth]{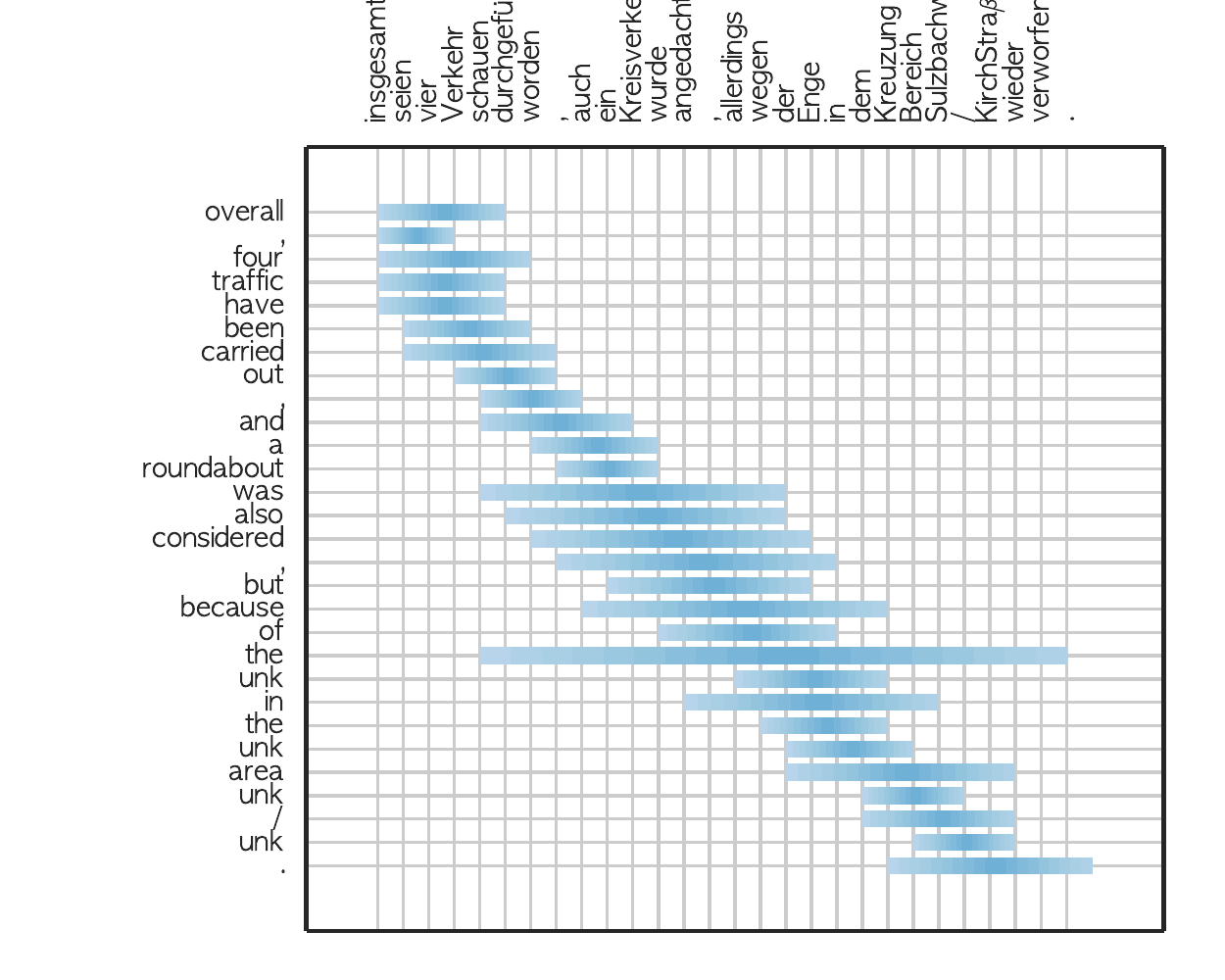}} \\
  \subfloat[]{\includegraphics[width = 0.5\textwidth]{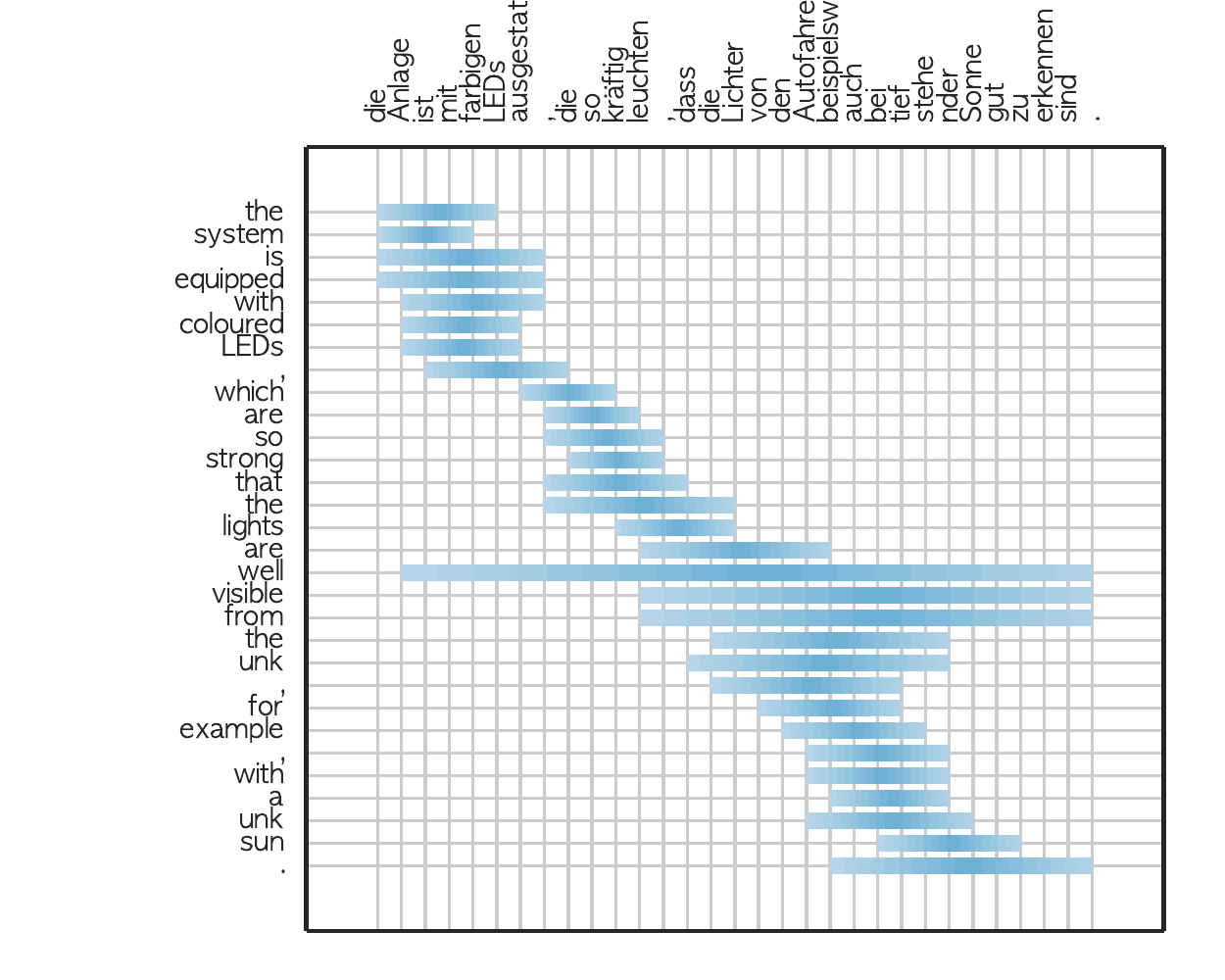}}
  \subfloat[]{\includegraphics[width = 0.5\textwidth]{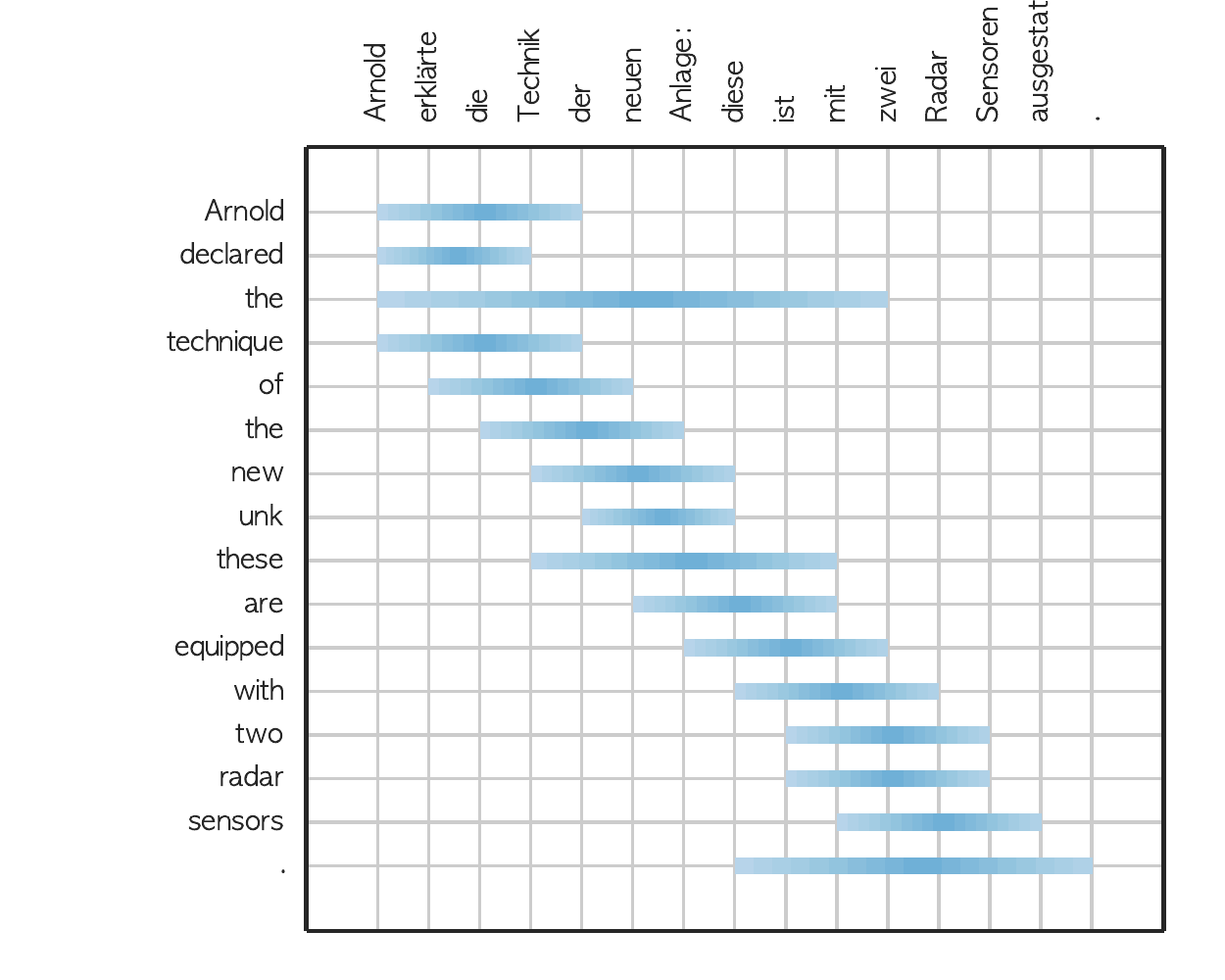}}
  \caption{A visionization of the vision spans predicted by Flexible Attention for six random long sentences in the De-En development corpus}
  \label{figure:random_vis}
\end{figure*}

\end{document}